# A Study on the Implementation Method of an Agent-Based Advanced RAG System Using Graph

**Cheonsu Jeong**[1]

[1] Dr. Jeong is Principal Consultant & the Technical Leader for AI Automation at SAMSUNG SDS;

**Abstract**

This study aims to improve knowledge-based question-answering (QA) systems by overcoming the limitations of existing Retrieval-Augmented Generation (RAG) models and implementing an advanced RAG system based on Graph technology to develop high-quality generative AI services. While existing RAG models demonstrate high accuracy and fluency by utilizing retrieved information, they may suffer from accuracy degradation as they generate responses using pre-loaded knowledge without reprocessing. Additionally, they cannot incorporate real-time data after the RAG configuration stage, leading to issues with contextual understanding and biased information.

To address these limitations, this study implemented an enhanced RAG system utilizing Graph technology. This system is designed to efficiently search and utilize information. Specifically, it employs LangGraph to evaluate the reliability of retrieved information and synthesizes diverse data to generate more accurate and enhanced responses. Furthermore, the study provides a detailed explanation of the system's operation, key implementation steps, and examples through implementation code and validation results, thereby enhancing the understanding of advanced RAG technology. This approach offers practical guidelines for implementing advanced RAG systems in corporate services, making it a valuable resource for practical application.

## I. Introduction

Recent advancements in AI technology have brought significant attention to Generative AI. Generative AI, a form of artificial intelligence that can create new content such as text, images, audio, and video based on vast amounts of trained data models (Jeong, 2023d), is being applied in various fields, including daily conversations, finance, healthcare, education, and entertainment (Ahn & Park, 2023). As generative AI services become more accessible to the general public, the role of generative AI-based chatbots is becoming increasingly important (Adam et al., 2021; Przegalinska et al., 2019; Park, 2024). A chatbot is an intelligent agent that allows users to have conversations typically through text or voice (Sánchez-Díaz et al., 2018; Jeong & Jeong, 2020). Recently, generative AI chatbots have advanced to the level of analyzing human emotions and intentions to provide responses (Jeong, 2023a). With the advent of large language models (LLMs), these chatbots can now be utilized for automatic dialogue generation and translation (Jeong, 2023b). However, they may generate responses that conflict with the latest information and have a low understanding of new problems or domains as they rely on previously trained data (Jeong, 2023c). While 2023 was marked by the release of foundational large language models (LLMs) like ChatGPT and Llama-2, experts predict that 2024 will be the year of Retrieval Augmented Generation (RAG) and AI Agents (Skelter Labs, 2024).

* Corresponding Author: Cheonsu Jeong; paripal@korea.ac.kr

However, there are several considerations for companies looking to adopt generative AI services. Companies must address concerns such as whether the AI can provide accurate responses based on internal data, the potential risk of internal data leakage, and how to integrate generative AI with corporate systems. Solutions include using domain-specific fine-tuned LLMs and enhancing reliability with RAG that utilizes internal information (Jung, 2024). When domain-specific information is fine-tuned on GPT-4 LLM, accuracy improves from 75% to 81%, and adding RAG can further increase accuracy to 86% (Angels et al., 2024). RAG models are known for effectively combining internal knowledge retrieval and generation to produce more accurate responses. They offer the advantages of source-based fact provision and addressing data freshness issues through the integration of internal and external knowledge bases. However, the effectiveness of RAG models heavily depends on the quality of the database, directly impacting model performance (Kim, 2024). Traditional RAG models load knowledge once and generate responses without reprocessing, leading to accuracy degradation and an inability to reflect real-time data after the RAG configuration. This process can result in inaccurate responses, particularly when generating answers to complex questions, as the initial vectorized knowledge is used without updating with new information. Furthermore, traditional RAG models struggle to handle various types of questions and may suffer from unrelated documents being used in response due to poor retrieval strategies, along with the hallucination issues observed in LLMs.

The purpose of this study is to improve the traditional RAG model-based knowledge-based QA system (Jeon et al., 2024) and overcome its limitations by accessing real-time data and verifying whether the retrieved documents are genuinely relevant to the questions. By implementing an enhanced RAG system capable of addressing questions about recent events and real-time data, and being less susceptible to hallucinations, this study aims to improve the quality and performance of generative AI services.

The introduction of this paper explains the research background and objectives, the limitations of existing RAG models, the importance and contributions of the study, and the structure of the paper. The theoretical background reviews the overview of RAG models, advanced RAG approaches, and case studies of existing research improvements. The design of the advanced RAG model covers the composition flow of advanced RAG, the configuration of Agent RAG, and other enhanced features. The implementation of the advanced RAG system details the overview and application of LangGraph, the system implementation process, and the results. The testing section presents the improved results of the implemented code. Finally, the conclusion summarizes the research findings, discusses the limitations, and outlines directions for future research.

# II. Related Work

For this study, recent key research papers, journals, and articles related to the RAG model were reviewed. This section provides an overview of the RAG model and describes the advancements leading to the development of the Advanced RAG.

## 2.1. Overview of the RAG Model

The RAG (Retrieval-Augmented Generation) model combines retrieval and generation to produce answers by integrating document retrieval and generation models (Lewis et al., 2020). To generate an answer to a question, the model first retrieves relevant documents and then uses them to produce the response. This process helps in generating accurate answers to questions. The RAG model can handle various types of questions effectively, even when there is a lack of specific domain knowledge. Consequently, it enhances the accuracy and consistency of information compared to traditional generative models.

The RAG model consists of two main stages:

- **Retrieval Stage:** Information relevant to the given question is retrieved through a search engine.
- **Generation Stage:** Answers are generated based on the retrieved information.

### 2.1.1. RAG Model Implementation Flow

The RAG model performs text generation tasks by retrieving information from a given source data and using that information to generate the desired text. The data processing for using RAG involves dividing the original data into smaller chunks, embedding the text data by converting it into numerical vectors, and storing these vectors in a vector store (Microsoft, 2023). The implementation flow of a generative AI service based on the RAG model is depicted in Figure 1 (Jeong, 2023e).

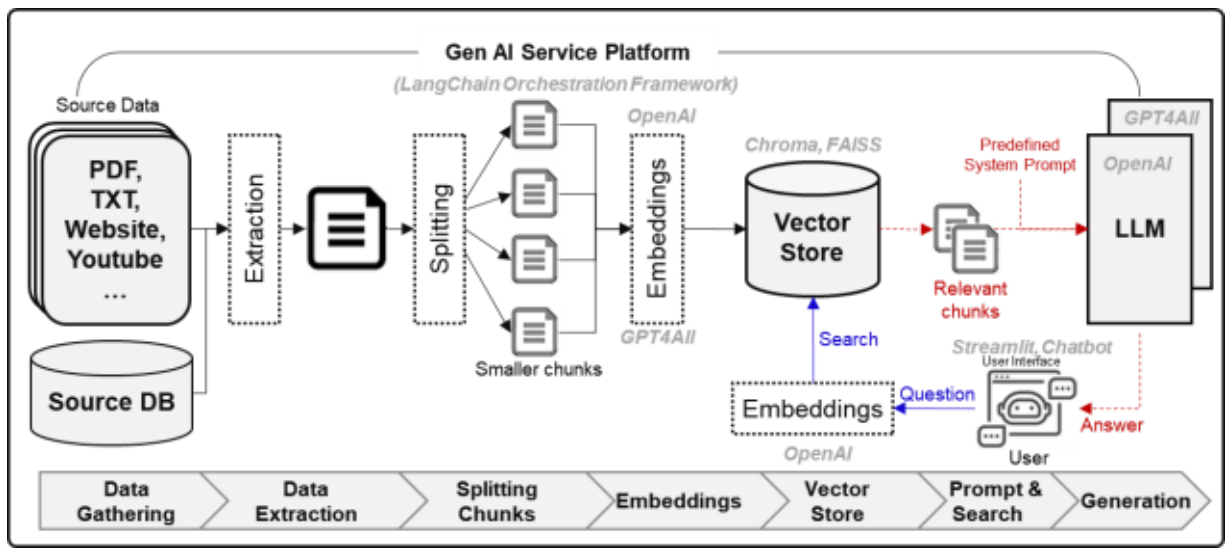

**Figure 1:** Implementation Flow of a RAG-based Generative AI Service

2.1.2. RAG-Based Vector Store Types

To establish a RAG (Retrieval-Augmented Generation) system, a vector database is utilized to store knowledge. A typical vector pipeline for vector databases involves three stages: Indexing, Querying, and Post Processing (Devtorium, 2023). Specifically, RAG-based vector store configurations can be categorized into two types as illustrated in Figure 2: one where all source data is pre-stored in the vector store and another where data is dynamically inserted at query time.

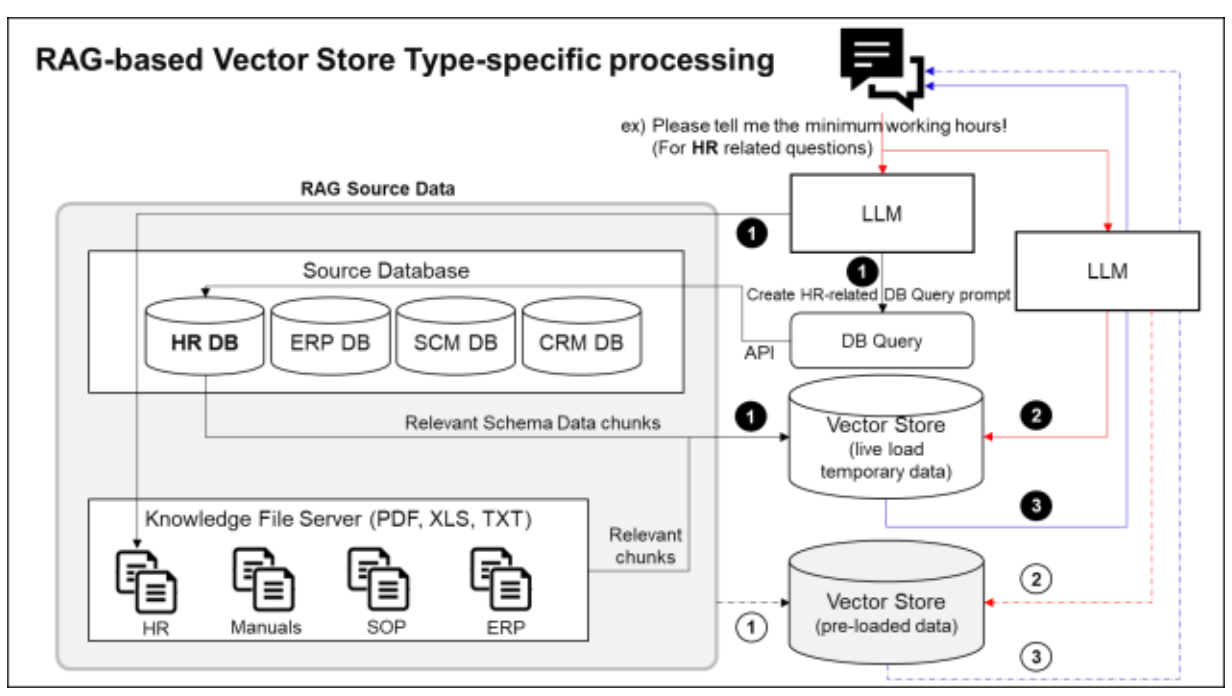

**Figure 2:** Configuration Types and Processing Procedures of RAG-Based Vector Stores

When a company offers internal knowledge through an Open LLM (Large Language Model), ensuring security is a critical issue, making the use of Local LLMs essential. In this scenario, it is effective to employ multiple Local LLMs, each optimized for different tasks. For instance, one LLM might be specialized in generating database queries to retrieve relevant data from multiple source databases, while another LLM could be developed to provide answers based on specific domain knowledge (Jeong, 2023e).

## 2.2. Prior Research on Advanced RAG

2.2.1. Methods to Enhance RAG Performance

The performance of RAG (Retrieval-Augmented Generation) is influenced by the quality of the data that can be composed into prompts based on the results of question processing from external repositories. Recently, various Advanced RAG (Advanced Retrieval-Augmented Generation) methods have been proposed to address the limitations of conventional RAG. Advanced RAG represents an evolved form of the traditional RAG technique, incorporating various optimization methods to overcome its limitations. Recent research by Yunfan G. et al. introduces an optimization strategy that divides the retrieval process into Pre-Retrieval, Retrieval, and Post-Retrieval stages, significantly enhancing information accuracy and processing efficiency through optimization at each stage (Yunfan G. et al., 2024). Additionally, various improvement methods, such as re-ranking based on relevance to enhance accuracy, have been proposed as strategies for improving the quality of RAG systems (Jang Dong-jin, 2024), as outlined in Table 1 (Matt A., 2023). Frameworks like LangChain or LlamaIndex provide libraries for implementing these strategies, making the implementation process more straightforward.

**Table 1:** Methods to Enhance RAG Performance

| Method | Descriptions |
|---|---|
| Clean your data | When dealing with conflicting or redundant information, it becomes challenging to find the correct context during retrieval. To ensure accurate responses to queries, it is essential to properly structure the documents themselves. One approach is to create summaries for all documents and use these summaries as context. |
| Explore different index types | While embedding-based similarity search methods generally perform well, they are not always the best solution. For example, in e-commerce, keyword-based search methods may be more suitable for finding specific items such as products. Many systems employ hybrid approaches, where keyword-based searches are used for specific products, and embedding-based searches are applied for general customer information and support. |
| Experiment with your chunking approach | Chunk size is critically important, with smaller chunks typically yielding better performance; however, they may also lead to issues related to insufficient surrounding context. Generally, smaller chunk sizes aid search systems in identifying relevant contextual information more effectively. |
| Play around with your base prompt | To reduce hallucinations, prompting should be designed to ensure that responses are based solely on the given contextual information. For example: "You are a customer support representative, designed to provide assistance based on factual information only. Please answer queries based on the given context information, not on pre-trained knowledge." |
| Try meta- | After adding relevant metadata tags to the |
| data filtering | chunks (such as document title, page number, email, date, etc.), these tags are used to process the results. |
| Use query routing | Having multiple indexes is often beneficial. When a query is received, it can then be routed to the appropriate index. For example, there may be one index for handling summary questions, another for addressing factual questions, and a third index suited for date-sensitive inquiries. |
| Look into reranking | Using re-ranking allows the search system to retrieve the top similar nodes based on context, and then re-rank them according to relevance, thereby enhancing accuracy. |
| Consider query transformations | If the relevant context for the initial question cannot be found, modifying the question and retrying can improve answer accuracy. This can be implemented in the RAG system by allowing the query to be decomposed into multiple questions. |
| Fine-tune your embedding model | When the context or domain does not align, fine-tuning the embedding model can enhance performance, particularly for domain-specific terminology. For example, this can involve adapting the model to better handle specialized vocabulary pertinent to a specific domain. |
| Start using LLM dev. tools | When building a RAG system using LlamaIndex or LangChain, debugging tools can be utilized to identify the sources of documents and context. |

2.2.2. Research on Advanced RAG Types

Notable advanced RAG approaches currently being researched include the following:

- **Self-RAG:** This method involves re-searching generated responses to find relevant information and using it to refine the answers. This approach can enhance the accuracy and fluency of the responses (Asai A. et al., 2023).
- **Corrective RAG:** This approach employs a Corrective Agent to rectify errors in generated responses. The Corrective Agent identifies errors in the responses and retrieves information to correct them, thereby improving the reliability of the answers (Yan, S.Q. et al., 2024).
- **Adaptive RAG:** This method involves selecting the appropriate RAG approach based on the type of question. For instance, Self-RAG may be used for factual questions, while Corrective RAG could be employed for opinion-based questions. By choosing the appropriate method according to the question type, the accuracy of the responses can be improved (Jeong, S. et al., 2024).

# III. Design of Advanced RAG Models

In this chapter, various Advanced RAG approaches proposed in previous research are reviewed, and an enhanced RAG system is designed based on these findings. Specifically, we closely analyze methods such as Self-RAG, Corrective RAG, and Adaptive RAG, and present an implementation model as shown in Figure 3 based on the improvements derived from these analyses. The implementation of the Agent RAG system primarily builds on Corrective RAG, while referencing Self-RAG and Adaptive RAG. The workflow to enhance a typical RAG system involves retrieving document chunks from a vector database and then using an LLM to verify the relevance of each retrieved document chunk to the input query. If all retrieved document chunks are relevant, the system proceeds with the standard RAG pipeline to generate a response using the LLM. However, if some retrieved documents are deemed irrelevant to the input query, the input query is rephrased, and the web is searched for new information relevant to the query. This new information is then sent to the LLM to generate a response. This workflow is implemented using a graph-based approach for the Advanced RAG system.

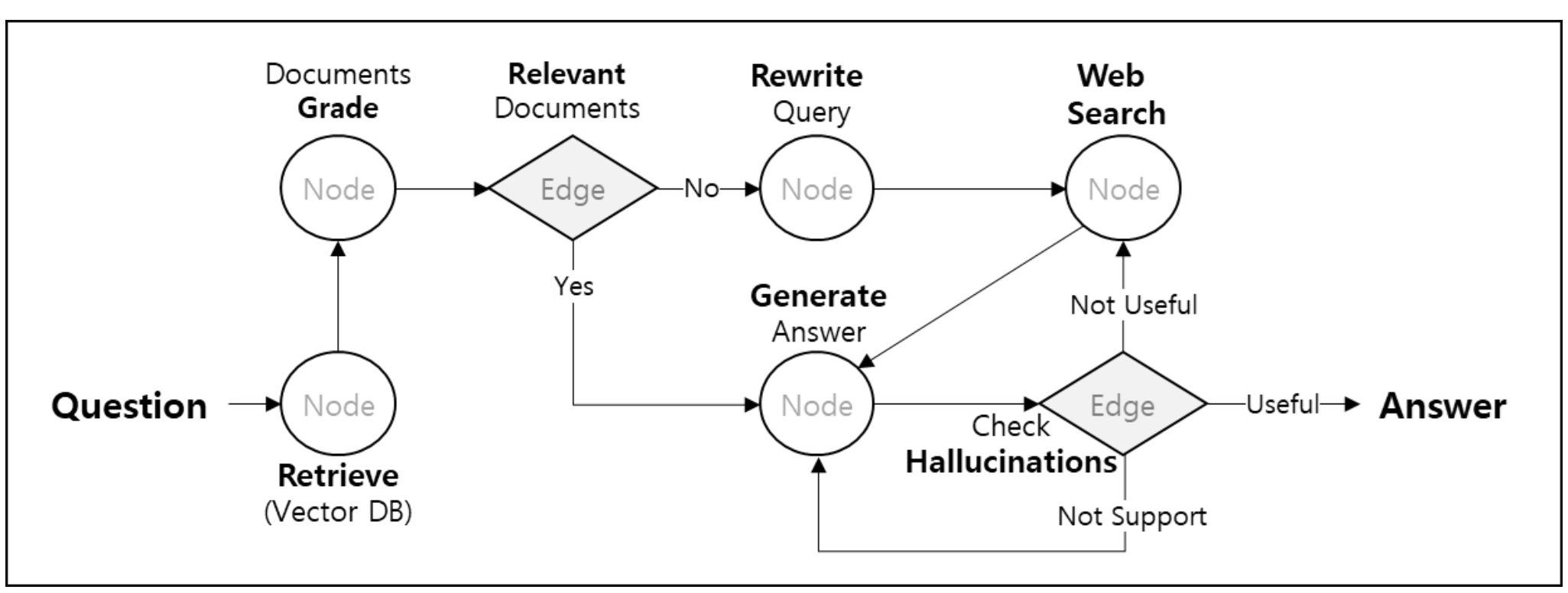

**Figure 3:** Agent-based advanced RAG workflow

### 3.1. Design of Advanced RAG Execution Procedures

The Advanced RAG model maintains the basic flow of a traditional RAG model while incorporating additional processes after the search stage to enhance the accuracy and consistency of responses. The composition and flow of the Advanced RAG model are designed as follows:

① **Query Processing:** Input the user's query, and analyze and understand the precise meaning of the query through intent recognition and analysis.

② **Search:** Utilize search engines to explore and retrieve information from various sources related to the query.

③ **Candidate Selection:** Select information from the search results that is highly relevant and reliable to the query as candidates.

④ **Candidate Ranking:** Rank the selected candidates based on their relevance to the query, the reliability of the information, and diversity.

⑤ **Answer Generation:** Use a text generation model to create an answer based on the ranked candidate information.

⑥ **Answer Updating:** Continuously collect new information and update the answer to provide the most current information.

The operation of the Advanced RAG model is as follows:

① Input the question.

② Search for information related to the question using a search engine.

③ Extract relevant information related to the question from the search results using an information extractor.

④ Generate an answer based on the extracted information.

⑤ The Agent refines the answer to enhance its accuracy and fluency.

⑥ Output the final answer.

### 3.2. Application of Agent RAG

To apply the enhanced execution procedures, the Agent RAG framework incorporates the concept of an "Agent" into the answer generation process, thereby further improving the accuracy and consistency of responses. The Agent serves as a core element in the answer generation process, fulfilling the following roles:

- **Answer Evaluation**: Assessing the accuracy, fluency, and reliability of the generated responses.
- **Answer Improvement**: Enhancing the responses based on the evaluation results.
- **Information Retrieval**: Searching for necessary information to improve the responses.

### 3.3. Application of the LangGraph Module

LangGraph is a module released by LangChain designed to build stateful multi-actor applications using LLMs. It is utilized to create Agent and multi-Agent workflows, allowing for the definition of flows that include essential cycles for most Agent architectures, and providing detailed control over the application's flow and state, which is critical for creating reliable Agents (LangGraph, 2024).

Built on top of LangChain, LangGraph facilitates the development of AI agents driven by LLMs by creating essential cyclic graphs. LangGraph treats Agent workflows as cyclic graph structures. Specifically, the LangGraph Conversational Retrieval Agent offers various functionalities essential for the development of language-based AI applications, including language processing, AI model integration, database management, and graph-based data processing. It is composed of states, nodes, and edges and performs the following roles:

- **Complex Workflow Management**: Useful for structuring and managing state-

based workflows, with clear transitions and branching for each stage.

- **Clear Flow Control**: Allows for precise definition of relationships between nodes and edges, facilitating the implementation of complex conditional logic and state transitions.

- **Scalability**: Enables easy addition of new nodes and edges to expand the workflow, accommodating complex systems with various conditional logic.

This study proposes using LangGraph, which offers a range of functionalities, as a suitable tool for implementing Agent-based Advanced RAG systems.

# IV. Implementation Results of the Advanced RAG System

This chapter presents the implementation of the RAG model and LangChain framework based on the Advanced RAG concepts introduced in Chapter 3, utilizing data suitable for internal corporate use. The implementation approach and considerations for using LangGraph, which is well-suited for Agent implementation, are demonstrated through practical examples.

## 4.1. Development Environment

The solutions and development platforms applied in this case are based on the framework outlined in Figure 1, and the implementation method utilizing LangGraph and OpenAI LLM is described. The process involves chunking and embedding documents, storing them in ChromaDB, and then transforming them into retrievers for document content search. The results are evaluated, and an Agent RAG Graph is defined and implemented accordingly. The development was carried out using Python, which provides a range of libraries necessary for AI development.

The development environment for each implementation component is as follows:

- Orchestration Framework: LangChain
- Agent Graph Workflow: LangGraph
- Workflow Trace: LangSmith
- Data Extraction and Chunking: LangChain Modules
- Embedding: OpenAI
- Vector Database: Chroma
- LLM: OpenAI GPT-4-turbo Model
- Python Development Environment: Google Colab

## 4.2. Results of the step-by-step implementation

### 4.2.1. Installation of Basic Libraries and API Key Setup

Basic libraries, including LangChain for overall orchestration of tasks such as data splitting, OpenAI for API access, ChromaDB for storing RAG knowledge, and a web search library, were installed. To facilitate easy management of knowledge files, Google Drive was integrated with Colab. To ensure security, the keys for various modules were registered in a .env file at a specific location. Figure4 shows that the .env file containing the keys was successfully read.

```
!pip install -q langchain_openai langchain_community langchainhub langchain langgraph

!pip install -q tiktoken chromadb pypdf tavily-python

!pip install -q python-dotenv  # .env file load

# Google Drive Connect
from google.colab import drive
drive.mount('/content/drive')

Mounted at /content/drive

cd /content/drive/MyDrive/RAG

/content/drive/MyDrive/RAG

# api key
from dotenv import load_dotenv # Add
load_dotenv() # load .env

True
```

**Figure 4:** Installation of Basic Libraries and OpenAI API Key Configuration

### 4.2.2. Retriever Implementation

For managing internal documents, such as 'Dress Code Standards.pdf'', the PyPDFLoader is used to load the document from the specified location. Given the document's characteristics, which include tables, the TextSplitter is adjusted. Instead of using the CharacterTextSplitter with a single delimiter (e.g., newline), the RecursiveCharacterTextSplitter is employed by adjusting the 'chunk_size' and 'chunk_overlap' parameters to efficiently maintain context. The split documents are stored in the Chroma vector store, embedded using OpenAIEmbeddings, and then converted into retrievers for search purposes.

```
from langchain.text_splitter import RecursiveCharacterTextSplitter
from langchain_openai import OpenAIEmbeddings
from langchain_community.vectorstores import Chroma
from langchain_openai import ChatOpenAI

llm = ChatOpenAI(model="gpt-4-turbo", temperature=0)
openai_embed_model = OpenAIEmbeddings(model='text-embedding-3-small')
```

```
from langchain.document_loaders import PyPDFLoader
loader = PyPDFLoader("/content/drive/MyDrive/RAGdata/Dress Code Standards.pdf")
docs = loader.load()

text_splitter = RecursiveCharacterTextSplitter(chunk_size=300, chunk_overlap=30)
chunked_docs = text_splitter.split_documents(docs)
chunked_docs[:2]
```

```
# Load the documents to vectorstore
vectorstore = Chroma.from_documents(documents=chunked_docs,collection_name="rag_pdf_db",
        embedding=openai_embed_model,)
retriever = vectorstore.as_retriever()
print(retriever)
```

```
tags=['Chroma', 'OpenAIEmbeddings'] vectorstore=<langchain_community.vectorstores.chroma.Chroma object
```

**Figure 5:** Data Loading and Embedding

### 4.2.3. Evaluation of Search Results

To assess whether the retrieved documents are relevant to the given question, the implementation is as shown in Figure 6.

```
### Implement the Retrieval Grader
from langchain_core.prompts import ChatPromptTemplate
from langchain_core.pydantic_v1 import BaseModel, Field
from langchain_openai import ChatOpenAI
# Data model for LLM output format
class GradeDocuments(BaseModel):
    """A binary score for checking relevance of retrieved documents."""
    binary_score: str = Field(
        description="Rates whether the document is relevant to the question with a 'yes' or 'no'
    )
# LLM for grading
llm = ChatOpenAI(model="gpt-4-turbo", temperature=0)
structured_llm_grader = llm.with_structured_output(GradeDocuments)
# Prompt template for grading
SYS_PROMPT = """You are a professional evaluator who evaluates the relevance of searched documen
                - We assign a relevance rating to documents if they contain keywords or semant
                - Your rating should be 'yes' or 'no' indicating whether the article is releva

grade_prompt = ChatPromptTemplate.from_messages(
    [
        ("system", SYS_PROMPT),
        ("human", """Searched Documents:{document}
                     User Questions:{question} """),
    ]
)
# Build grader chain
doc_grader = (grade_prompt | structured_llm_grader)
print(doc_grader)
```

```
first=ChatPromptTemplate(input_variables=['document', 'question'], messages=[SystemMessagePromptTemplate(prompt
```

**Figure 6:** Implementation of Search Results Evaluation

The retrieved documents are linked to user questions. If the document contains keywords related to the user's query, it is assessed for relevance with the aim of filtering out irrelevant searches. As depicted in Figure 7, relevance is indicated by assigning a binary score of 'yes' or 'no' (e.g., "GRADE: binary_score='yes'").

```
query = "What are some considerations for dress selection for work?"
topk_docs = similarity_threshold_retriever.invoke(query)
for doc in topk_docs:
    print(doc.page_content)
    print("================================")
    print('GRADE:', doc_grader.invoke({"question": query,
                                        "document": doc.page_content}))
    print("================================")
```

```
Dress Code Guidelines for Work
1. Purpose
   Taking into account the individuality of each employee according to the nature of the business, employe
expected to wear work attire that reflects professionalism and decorum.
2. Considerations for Dress Selection
================================
GRADE: binary_score='yes'
================================
(1) All employees.
   (2) For employees working with client companies, follow the dress code of the respective client.
4. Dress Standards
(1) Maintain a neat and simple  business casual attire, with the option of jeans and sneakers.
================================
GRADE: binary_score='yes'
================================
```

**Figure 7:** Relevance Determination of Search Results

Subsequently, a question-answer RAG chain is constructed, similar to traditional RAG systems, to integrate with the AI agent as shown in Figure 8.

```
# Build QA_RAG Chain
from langchain_core.prompts import ChatPromptTemplate
from langchain_openai import ChatOpenAI
from langchain_core.runnables import RunnablePassthrough, RunnableLambda
from langchain_core.output_parsers import StrOutputParser
from operator import itemgetter
# RAG prompt for generating answer
prompt = """You are an Assistant for a Q&A.
Answer the question using the following retrieved Context fragment.
If there is no context or you do not know the answer, answer that you do not know the answer
Do not construct an answer unless it corresponds to the provided Context.
If the Context value is null when constructing an answer,
  answer 'RAG does not have relevant information'.
Please provide a detailed and summarized answer to the question.
            Question: {question}
            Context: {context}
            Answer:
          """
prompt_template = ChatPromptTemplate.from_template(prompt)
# Initialize GPT-4-turbo connection
llm = ChatOpenAI(model_name='gpt-4-turbo', temperature=0)
# Used to separate context documents on a new line
def format_docs(docs):
    return "₩n₩n".join(doc.page_content for doc in docs)
# Create QA RAG chain
qa_rag_chain = (
    {
        "context": (itemgetter('context') | RunnableLambda(format_docs)),
        "question": itemgetter('question')
    } | prompt_template | llm | StrOutputParser())
```

**Figure 8:** Question-Answer RAG Chain

Using the constructed question-answer RAG chain, the relevant answer results are obtained as illustrated in Figure 9.

```
query = "What are some considerations for dress selection for work?"
topk_docs = similarity_threshold_retriever.invoke(query)
result = qa_rag_chain.invoke(
    {"context": topk_docs, "question": query}
)
print(result)
```

```
Some considerations for dress selection for work include:

1. Reflecting professionalism and decorum in attire according to the nature of the business.
2. For employees interacting with client companies, adhering to the dress code of the respective client.
3. Maintaining a neat and simple business casual attire, which may include jeans and sneakers depending
```

**Figure 9:** Answer Processing for Related Questions

However, as shown in Figure 10, when an out-of-context question is attempted, the response "RAG does not have relevant information" is received, indicating that the question cannot be answered.

```
query = "Tell me what is the capital of the country where BTS is located"
topk_docs = similarity_threshold_retriever.invoke(query)
result = qa_rag_chain.invoke(
        {"context": topk_docs, "question": query}
)
print(result)
```

```
/usr/local/lib/python3.10/dist-packages/langchain_core/vectorstores/base.py:784: UserWarning: Relevance scores
  warnings.warn(
/usr/local/lib/python3.10/dist-packages/langchain_core/vectorstores/base.py:796: UserWarning: No relevant docs
  warnings.warn(
RAG does not have relevant information.
```

**Figure 10:** Answer Processing for Unrelated Questions

To improve this, the question can be rephrased into a more optimized version for web search, as shown in Figure 11. Rewriting the question (question rewriting) helps obtain better contextual information from the web. For example, an incomplete question like "Tell me what is the capital of the country where BTS is located" can be improved to "What is the capital of South Korea, the country where BTS is from?"

```
# Question rewriting - LLM : Rewrites entered user queries into questions optimized for web search
llm = ChatOpenAI(model="gpt-4-turbo", temperature=0)
# Question rewriting - Prompt template
SYS_PROMPT = """Act as a question rewriter and perform the following tasks:
                - Convert the following input question into a better version optimized for web searche
                - When rewriting, look at the input question and infer its underlying semantic intent,
             """
re_write_prompt = ChatPromptTemplate.from_messages(
    [
        ("system", SYS_PROMPT),
        ("human", """First Questions:
                    {question}
                    Write improved questions.
                  """,
        ),
    ]
)
# Create a rephraser chain
question_rewriter = (re_write_prompt| llm | StrOutputParser())
```

```
query = "Tell me what is the capital of the country where BTS is located"
question_rewriter.invoke({"question": query})
```

```
'What is the capital of South Korea, the country where BTS is from?'
```

**Figure 11:** Question Rewriting and Improved Question

Additionally, the generated answers are evaluated to determine if they contain hallucinations, as depicted in Figure 12.

```
### Implement the hallucination grader
from langchain.prompts import PromptTemplate
from langchain_core.output_parsers import JsonOutputParser
prompt = PromptTemplate(
    template="""  This is an evaluator that evaluates whether or not a hallucination is present.
    It gives a binary 'yes' or 'no' to indicate whether or not a hallucination is present.
    If you have hallucinations, give 'yes'.
  Provide the binary score as a JSON with a
    single key 'score' and no preamble or explanation.
    Here are the facts:
    ₩n ------- ₩n
    {documents}
    ₩n ------- ₩n
    Here is the answer: {generation}  """,
    input_variables=["generation", "documents"],
)
hallucination_grader = prompt | llm | JsonOutputParser()

# Run
generation = rag_chain.invoke({"context": docs, "question": query})
hallucination_grader_response = hallucination_grader.invoke({"documents": docs, "generation": generation})
print(hallucination_grader_response)
```

```
{'score': 'no'}
```

**Figure 12:** Evaluation of Hallucinations

Subsequently, the answers are assessed for their utility in solving the question, as shown in Figure 13. A binary score of 'yes' or 'no' is assigned to indicate whether the answer is useful for solving the question.

```
### Implement the Retrieval Grader : Evaluate whether the document contains keywords relevant to the user's questi
prompt = PromptTemplate(
    template=""" You are a relevance evaluator.
You connect the retrieved document to the user's question. If the document contains keywords relevant to the user'
rate it as relevant. There is no need to be strict. The goal is to filter out bad searches. ₩n
You assign a binary score of 'yes' or 'no' to indicate whether the document is relevant to the question. ₩n
    Provide the binary score as a JSON with a single key 'score' and no premable or explaination.
    ₩n ------- ₩n
    {documents}
    ₩n ------- ₩n
    Here is the answer: {generation}""",
    input_variables=["generation", "documents"],
)
retrieval_grader = prompt | llm | JsonOutputParser()

# Run
result = rag_chain.invoke({"context": docs, "question": query})
retrieval_grader_response = retrieval_grader.invoke({"documents": docs, "generation": result})
print(retrieval_grader_response)
```

```
{'score': 'no'}
```

**Figure 13:** Evaluation of Answer Relevance

### 4.2.4. Definition of the Agent RAG Graph

To enhance answer retrieval, the Tavily API is used for web searches, and the connection to this API is established. The Graph State of the Agent is defined, where the state object is passed to each node in the graph. Nodes such as Retrieve, generate_answer, grade_documents, and web_search_add are defined as shown in Figure 14.

```
# Load web search tool
from langchain_community.tools.tavily_search import TavilySearchResults
web_search_tool = TavilySearchResults(max_results=2, search_depth='advanced',max_tokens=5000)
```

```
#Define the Graph State
from typing_extensions import TypedDict
from typing import List

### State
class GraphState(TypedDict):
    """
    The state of the graph.
    Attributes:
        question: question
        generation: Generate LLM Answer
        web_search_add: Flag indicating whether to add web search - yes or no
        documents: List of context documents
    """
    question : str
    generation : str
    web_search_add : str
    documents : List[str]
```

**Figure 14:** Definition of Web Search Tool and Agent Graph State

```
######## Node definition ########
# Search for documents in the vector store
from langchain.schema import Document
def retrieve(state):
    """
    Search for documents in the vector store
    Args:
        state (dict): Current graph status
    Returns:
        state (dict): Add new keys to documents, including those that contain the searched documents
    """
    print("---Search in VectorDB---")
    question = state["question"]
    # Retrieval
    documents = similarity_threshold_retriever.invoke(question)
    return {"documents": documents, "question": question}
```

**Figure 15:** Example of Retrieve Node Graph Implementation

The Agent RAG Graph can be composed of nodes such as Retrieve, grade_documents, rewrite_query, web_search_add, and generate_answer, as illustrated in Figure 14. The State, consisting of a set of messages, is used to store and represent the state of the agent graph as it passes through various nodes. Figure 15 shows the implementation example of the Retrieve Node Graph, which is used to fetch relevant contextual documents from the vector database. It also defines the node classes for document evaluation (grade_documents), question rewriting (rewrite_query), web searching (web_search_add), and answer generation (generate_answer).

#### 4.2.5. Implementation of the Agent RAG Graph

In the implementation phase of the Agent RAG Graph, LangGraph is used to build the Agent into a graph by utilizing the functions developed in the previous section. This involves placing the Agent into relevant nodes and connecting them with defined edges according to the specified workflow. The Agent performs an action that calls the Retrieve function and then adds output information to the state before invoking the Agent. As shown in Figure 16, the StateGraph class is used to define and manage the state-based graph. The provided code sets up the workflow to define the process for retrieving documents or performing other tasks based on the Agent's decisions.

```python
from langgraph.graph import END, StateGraph
workflow_agent_rag = StateGraph(GraphState)
# Define the nodes
workflow_agent_rag.add_node("retrieve", retrieve)  # retrieve
workflow_agent_rag.add_node("grade_documents", grade_documents)  # grade documents
workflow_agent_rag.add_node("rewrite_query", rewrite_query)  # transform_query
workflow_agent_rag.add_node("web_search", web_search)  # web search
workflow_agent_rag.add_node("generate_answer", generate_answer)  # generate answer
# Build graph
workflow_agent_rag.set_entry_point("retrieve")
workflow_agent_rag.add_edge("retrieve", "grade_documents")
workflow_agent_rag.add_conditional_edges("grade_documents",decide_to_generate,
        {"rewrite_query": "rewrite_query", "generate_answer": "generate_answer"},)
workflow_agent_rag.add_edge("rewrite_query", "web_search")
workflow_agent_rag.add_edge("web_search", "generate_answer")
workflow_agent_rag.add_conditional_edges(
        "generate_answer",
        grade_generation_v_documents_and_question,
        {
            "not supported": "generate_answer",
            "useful": END,
            "not useful": "web_search",
        },
)
# Compile
workflow_agent_rag = workflow_agent_rag.compile()
```

**Figure 16:** Node Definition and Graph Creation for Answer Generation

After the retrieve node, document evaluation (grade_documents) is performed to determine whether to proceed with generate or web_search tasks. This includes completing the tasks along each path or returning to the Agent. This process is used to manage complex decisions and task flows, ultimately compiling the workflow to create an executable application. The generated workflow can be visualized using IPython.display, as shown in Figure 17.

```python
from IPython.display import Image, display, Markdown
display(Image(workflow_agent_rag.get_graph().draw_mermaid_png()))
```

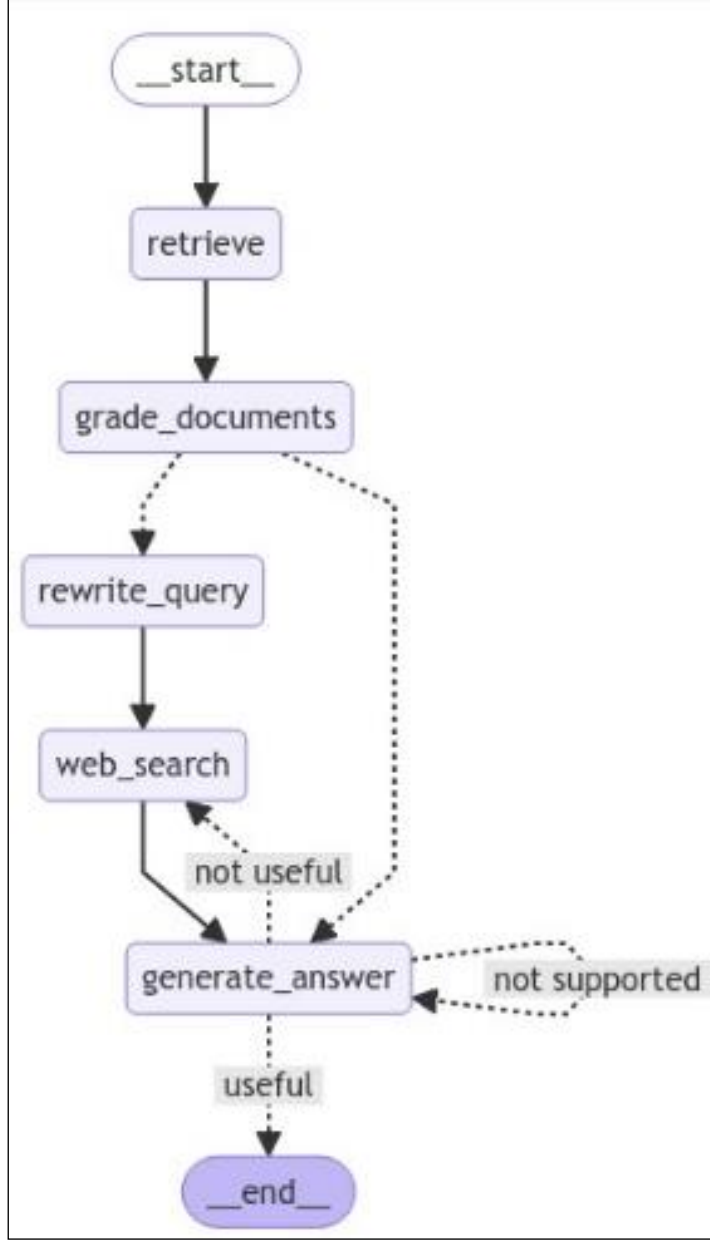

**Figure 17:** Graph of the Answer Generation Process

### 4.3. Test Results

The implemented Agent RAG workflow was tested with various questions to improve the accuracy of the answers, and the process of streaming responses to questions can be confirmed through the stream method.

#### 4.3.1. Verification of Questions in RAG Knowledge Information

When the question "Tell me the things to consider when choosing a work uniform" was input for the document ‘Dress Code Standards’ Figure 18, which is RAG information, the question was rewritten to "What are the main factors to consider when choosing a work uniform?" to improve the accuracy of the answer. This resulted in a more accurate response. This process can be confirmed through the streaming of responses to questions using the stream method as shown in Figure 19.

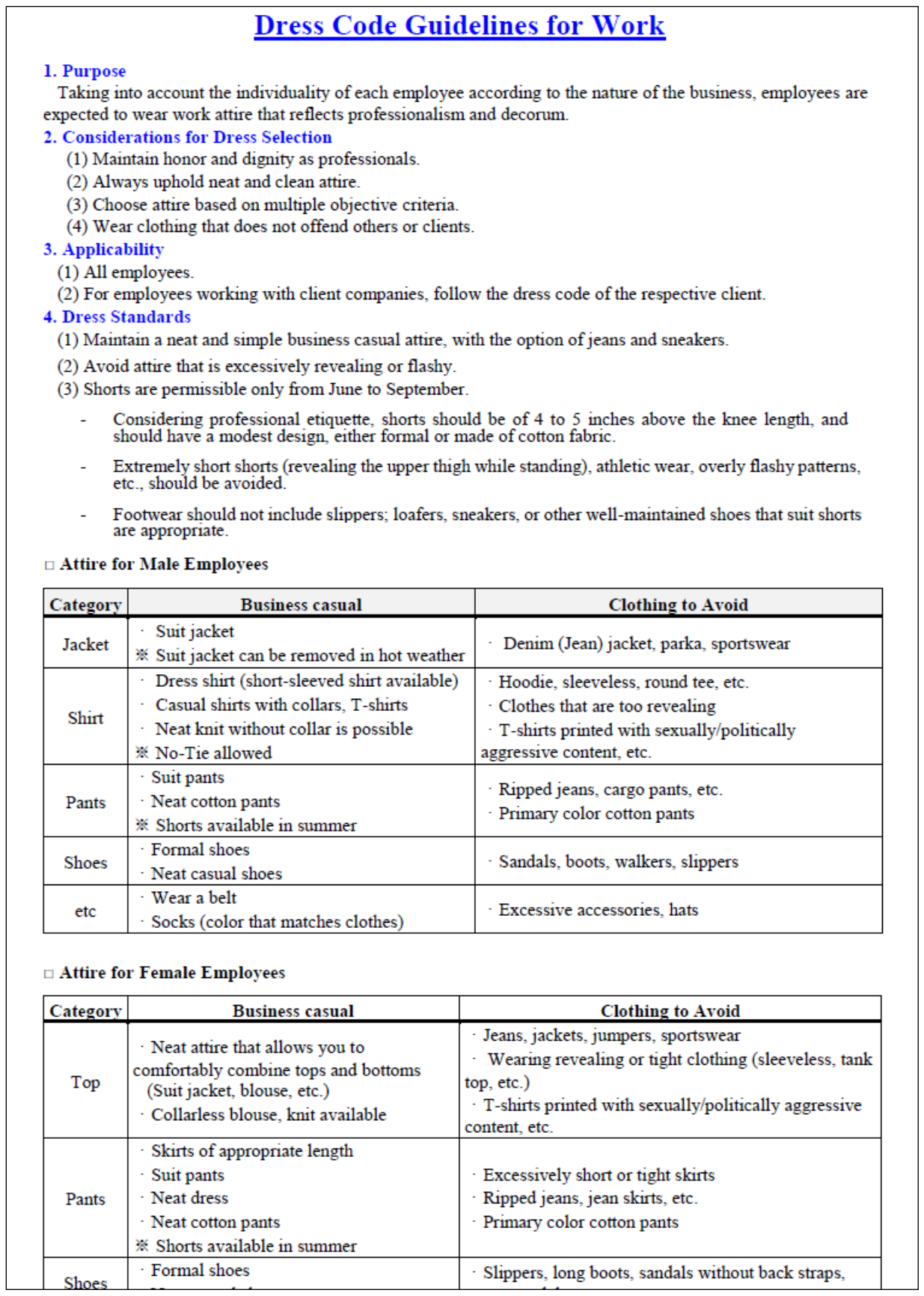

## Dress Code Guidelines for Work

**1. Purpose**
Taking into account the individuality of each employee according to the nature of the business, employees are expected to wear work attire that reflects professionalism and decorum.

**2. Considerations for Dress Selection**
(1) Maintain honor and dignity as professionals.
(2) Always uphold neat and clean attire.
(3) Choose attire based on multiple objective criteria.
(4) Wear clothing that does not offend others or clients.

**3. Applicability**
(1) All employees.
(2) For employees working with client companies, follow the dress code of the respective client.

**4. Dress Standards**
(1) Maintain a neat and simple business casual attire, with the option of jeans and sneakers.
(2) Avoid attire that is excessively revealing or flashy.
(3) Shorts are permissible only from June to September.

- Considering professional etiquette, shorts should be of 4 to 5 inches above the knee length, and should have a modest design, either formal or made of cotton fabric.
- Extremely short shorts (revealing the upper thigh while standing), athletic wear, overly flashy patterns, etc., should be avoided.
- Footwear should not include slippers; loafers, sneakers, or other well-maintained shoes that suit shorts are appropriate.

□ **Attire for Male Employees**

| Category | Business casual | Clothing to Avoid |
|---|---|---|
| Jacket | · Suit jacket<br>※ Suit jacket can be removed in hot weather | · Denim (Jean) jacket, parka, sportswear |
| Shirt | · Dress shirt (short-sleeved shirt available)<br>· Casual shirts with collars, T-shirts<br>· Neat knit without collar is possible<br>※ No-Tie allowed | · Hoodie, sleeveless, round tee, etc.<br>· Clothes that are too revealing<br>· T-shirts printed with sexually/politically aggressive content, etc. |
| Pants | · Suit pants<br>· Neat cotton pants<br>※ Shorts available in summer | · Ripped jeans, cargo pants, etc.<br>· Primary color cotton pants |
| Shoes | · Formal shoes<br>· Neat casual shoes | · Sandals, boots, walkers, slippers |
| etc | · Wear a belt<br>· Socks (color that matches clothes) | · Excessive accessories, hats |

□ **Attire for Female Employees**

| Category | Business casual | Clothing to Avoid |
|---|---|---|
| Top | · Neat attire that allows you to comfortably combine tops and bottoms (Suit jacket, blouse, etc.)<br>· Collarless blouse, knit available | · Jeans, jackets, jumpers, sportswear<br>· Wearing revealing or tight clothing (sleeveless, tank top, etc.)<br>· T-shirts printed with sexually/politically aggressive content, etc. |
| Pants | · Skirts of appropriate length<br>· Suit pants<br>· Neat dress<br>· Neat cotton pants<br>※ Shorts available in summer | · Excessively short or tight skirts<br>· Ripped jeans, jean skirts, etc.<br>· Primary color cotton pants |
| Shoes | · Formal shoes | · Slippers, long boots, sandals without back straps, |

**Figure 18:** Dress Code Standards.pdf

```
query = "What are some considerations for dress selection for work?"
response = workflow_agent_rag.invoke({"question": query})
print("* First question : "+ query)
print("* Rewritten question : "+question_rewriter.invoke({"question": query}))
display(Markdown(response['generation']))
```

```
---Search in VectorDB---
---Check the relevance of questions and documents---
---GRADE: This is a relevant document---
---GRADE: This is a relevant document---
---Evaluate graded documents---
---DECISION: Generate response---
---Generate Answer---
---CHECK HALLUCINATIONS---
---DECISION: Generate document-based answers
---GRADE GENERATION vs QUESTION---
---DECISION: Generate answers to your questions---
* First question : What are some considerations for dress selection for work?
* Rewritten question : How should one choose appropriate attire for a professional setting?
```

Some considerations for dress selection for work include:

1. Reflecting professionalism and decorum in attire according to the nature of the business.
2. For employees interacting with client companies, adhering to the dress code of the respective client.
3. Maintaining a neat and simple business casual attire, which may include jeans and sneakers depending on the workplace standards.

**Figure 19:** Answer Generation Process for Information in RAG Knowledge

### 4.3.2. Verification of Questions Not in RAG Knowledge Information

When the question "Tell me what is the capital of the country where BTS is located" was input, which is not in the RAG knowledge information, it was determined to "perform a web search because all Vector RAG documents are irrelevant to the question." The incomplete initial question "Tell me what is the capital of the country where BTS is located" was rewritten to a complete and rewritten question " What is the capital of South Korea, the country where BTS is from?" and the query was made, resulting in the correct answer "The capital of South Korea is Seoul." This process can be confirmed through the streaming process as shown in Figure 20.

```
query = "Tell me what is the capital of the country where BTS is located"
response = workflow_agent_rag.invoke({"question": query})
print("* First question : "+ query)
print("* Rewritten question : "+question_rewriter.invoke({"question": query}))
display(Markdown(response['generation']))
```

```
---Search in VectorDB---
/usr/local/lib/python3.10/dist-packages/langchain_core/vectorstores/base.py:784: UserWarning: Relevance scores
  warnings.warn(
/usr/local/lib/python3.10/dist-packages/langchain_core/vectorstores/base.py:796: UserWarning: No relevant docs
  warnings.warn(
---Check the relevance of questions and documents---
---No documents found---
---Evaluate graded documents---
---DECISION: The Vectore RAG document is not relevant to the question, so I decided to search the web.---
---Query rewrite---
---WEB SEARCH---
---Generate Answer---
---CHECK HALLUCINATIONS---
---DECISION: Generate document-based answers
---GRADE GENERATION vs QUESTION---
---DECISION: Generate answers to your questions---
* First question : Tell me what is the capital of the country where BTS is located
* Rewritten question : What is the capital of South Korea, the country where BTS is from?
```

The capital of South Korea is Seoul.

**Figure 20** Answer Generation Process through Web Search for Information Not in RAG Knowledge

As shown in Figure 21 by tracing the execution using LangSmith, it can be confirmed that the process of generating answers through web search for information not in the RAG knowledge base proceeded according to the designed Agent workflow (retrieve, grade_documents, rewrite_query, web_search, generate_answer).

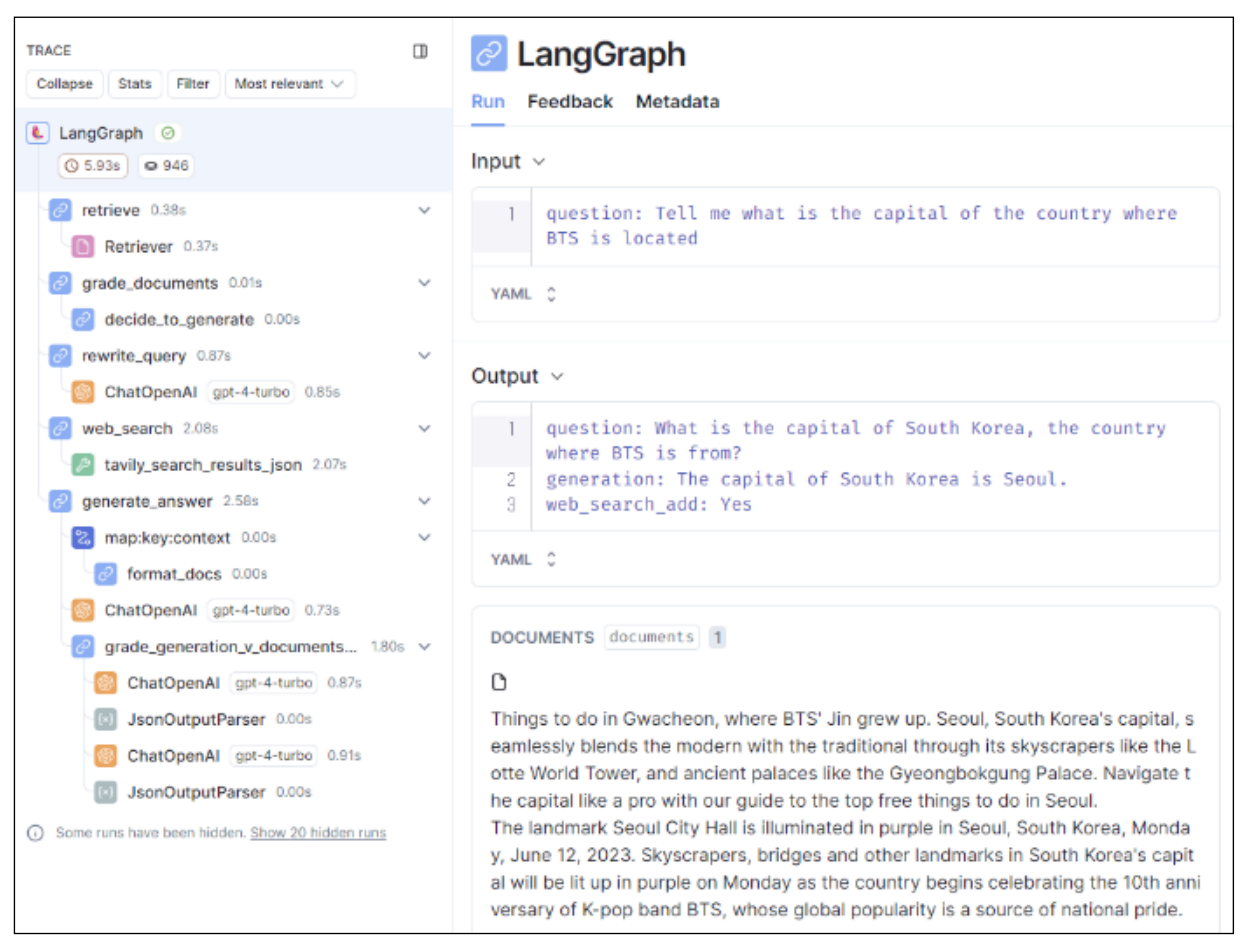

**Figure 21** Workflow Trace via LangSmith

# V. Conclusion and Discussion

This study has reviewed various methods to enhance the accuracy of RAG and explored the theoretical background of Advanced RAG models aimed at improving knowledge-based QA systems. Through the implementation of a graph-based Agent RAG system, along with specific implementation code and validation results, this research has demonstrated the feasibility of an enhanced RAG system.

The proposed graph-based Advanced RAG system offers a novel approach that significantly improves RAG performance, addressing the limitations of existing RAG models. The experimental results show that this system markedly enhances the accuracy and relevance of responses to user queries. The utilization of LangGraph's graph technology has effectively assessed the reliability of information, contributing to the improvement of information quality through question rewriting and web search optimization. By enhancing real-time data accessibility and strengthening the system's ability to handle various types of questions, the LangGraph-based method has expanded the potential applications of AI-driven customer support and information retrieval. These findings provide a crucial foundation for the advancement of RAG-based generative AI services.

However, several limitations remain. The LangGraph-based system is optimized for specific domains, which may result in performance degradation when applied to other fields. Additionally, the system's complexity may require additional resources for implementation and maintenance. Further validation processes are necessary to ensure the accuracy and reliability of real-time data, which could impact overall system performance.

Future research should focus on improving the generalizability of graph-based RAG systems. Expanding the system's applicability through testing and optimization across various domains, as well as developing and validating algorithms to enhance real-time data reliability, will be essential for further performance improvements. Lastly, given the rapid advancements in RAG technology, it is crucial to not only keep pace with technological progress but also to deeply understand and continually improve how information is retrieved and how accurate and reliable answers are generated.